\documentclass[sigconf]{acmart}

\usepackage{color}
\usepackage{booktabs}
\usepackage{multirow}
\usepackage{arydshln}

\usepackage{pifont}
\usepackage{bm}
\usepackage{colortbl}
\definecolor{mypink}{rgb}{.99,.91,.95}
\definecolor{mygray}{gray}{.9}
\AtBeginDocument{%
  \providecommand\BibTeX{{%
    \normalfont B\kern-0.5em{\scshape i\kern-0.25em b}\kern-0.8em\TeX}}}

\setcopyright{acmcopyright}
\copyrightyear{2022}
\acmYear{2022}
\setcopyright{acmcopyright}
\acmConference[MM '22]{Proceedings of the 30th ACM International Conference on Multimedia}{October 10--14, 2022}{Lisboa, Portugal}
\acmBooktitle{Proceedings of the 30th ACM International Conference on Multimedia
(MM '22), October 10--14, 2022, Lisboa, Portugal}
\acmPrice{15.00}
\acmDOI{10.1145/3503161.3547764}
\acmISBN{978-1-4503-9203-7/22/10}
\begin{document}

\title{Dynamic Prototype Mask for Occluded Person Re-Identification}
\author{Lei Tan$^{1}$, Pingyang Dai$^{1*}$, Rongrong Ji$^{1,2}$, Yongjian Wu$^{3}$}
\email{tanlei@stu.xmu.edu.cn,{pydai,rrji}@xmu.edu.cn, littlekenwu@tencent.com}
\affiliation{
    \institution{
    $^{1}$Media Analytics and Computing Lab, Department of Artificial Intelligence, School of Informatics,\\ Xiamen University, 361005, China, 
    $^{2}$Institute of Artificial Intelligence, Xiamen University, China\\ 
    $^{3}$Tencent Youtu Lab, Shanghai, China
    \country{}
    }
}
\def\authors{Lei Tan, Pingyang Dai, Rongrong Ji, Yongjian Wu}
\renewcommand{\shortauthors}{Tan et al.}


\begin{abstract}
  Although person re-identification has achieved an impressive improvement in recent years, the common occlusion case caused by different obstacles is still an unsettled issue in real application scenarios. Existing methods mainly address this issue by employing body clues provided by an extra network to distinguish the visible part. Nevertheless, the inevitable domain gap between the assistant model and the ReID datasets has highly increased the difficulty to obtain an effective and efficient model. To escape from the extra pre-trained networks and achieve an automatic alignment in an end-to-end trainable network, we propose a novel Dynamic Prototype Mask (DPM) based on two self-evident prior knowledge. Specifically, we first devise a Hierarchical Mask Generator which utilizes the hierarchical semantic to select the visible pattern space between the high-quality holistic prototype and the feature representation of the occluded input image. Under this condition, the occluded representation could be well aligned in a selected subspace spontaneously. Then, to enrich the feature representation of the high-quality holistic prototype and provide a more complete feature space, we introduce a Head Enrich Module to encourage different heads to aggregate different patterns representation in the whole image. Extensive experimental evaluations conducted on occluded and holistic person re-identification benchmarks demonstrate the superior performance of the DPM over the state-of-the-art methods. The code is released at \textcolor{magenta}{\url{https://github.com/stone96123/DPM}}.
  \renewcommand{\thefootnote}{}
  \footnotetext{* Corresponding Author.}
\end{abstract}

%





\maketitle

\section{Introduction}
\label{Sec:intro}
Person re-identification (ReID), aiming to address the problem of matching people over a distributed set of non-overlapping cameras, has attracted intensive attention in the last few years due to its wide applications in surveillance systems \cite{eom2019learning,Zhai2020ad,zheng2019pyramidal,ye2021channel}. While recent large-scale re-id datasets~\cite{zheng2017unlabeled,wei2018person} have provided an ability for deep neural networks to produce a satisfying retrieval performance upon the holistic pedestrian regions, widespread occlusion caused by different obstacles is still an unsettled issue in real application scenarios. This condition in practice inspires a large amount of research effort to explore the occluded person re-identification. 

\begin{figure}[t]
    \centering
    \includegraphics[width=0.9\columnwidth]{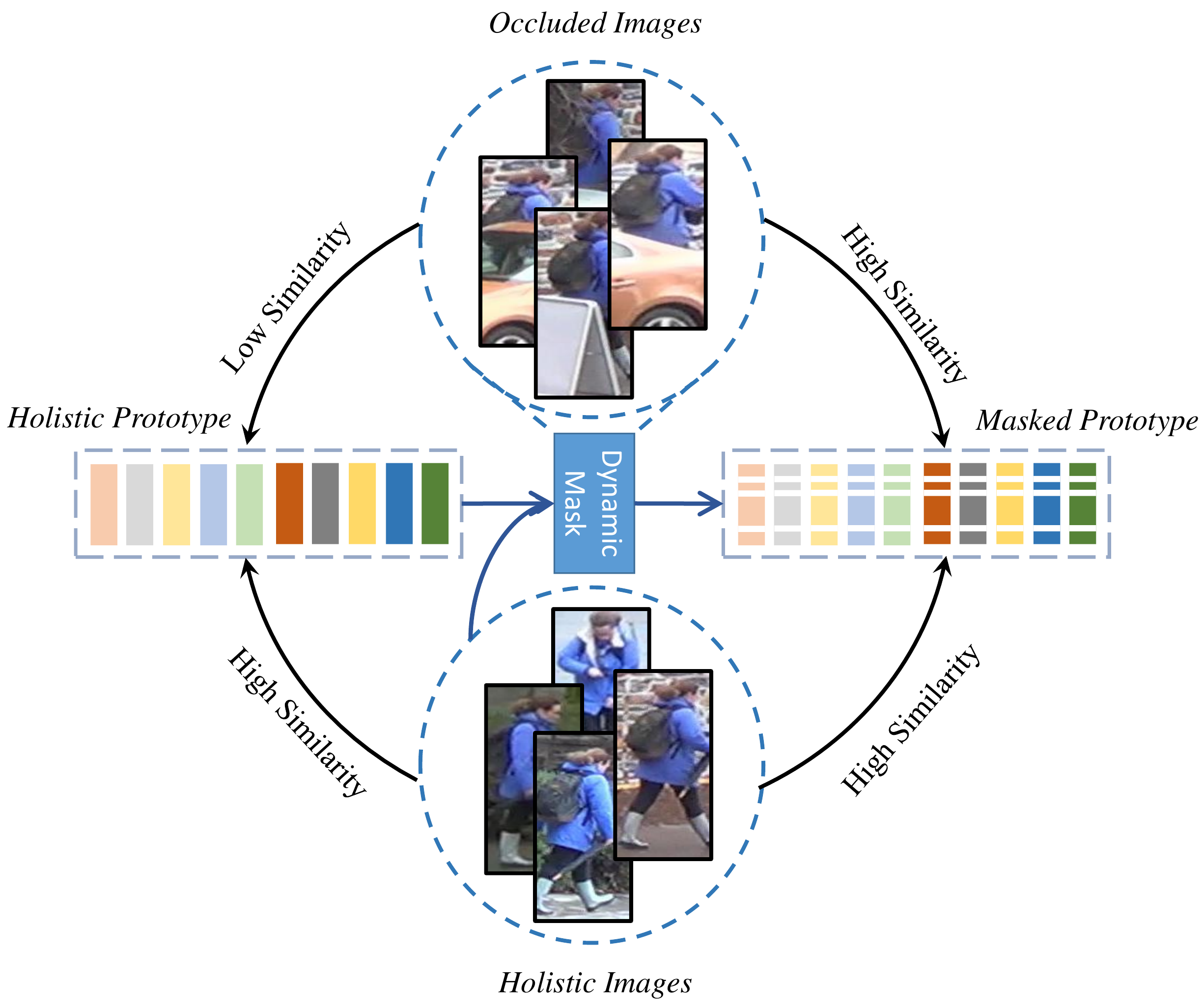}
    \caption{\textbf{The motivation for Dynamic Prototype Mask (DPM).} Based on two prior knowledge(detail discussed in Section~\ref{Sec:intro}), we consider the prototype as an ideal holistic representation for each class. Since the occluded image can not provide a complete representation, under DPM, the problem of alignment will be changed to find a visible pattern subspace from the holistic prototype.}
    \label{fig:softmax}
\end{figure}

Compared with the general person re-identification problem which assumes the whole body part is available,  the main challenge of occluded person re-identification is two folds: 
Firstly, with the obstacles which cover those discriminative body regions, the valid information in the final feature representation will high-decreased. Even if those valid regions provide valuable information, the final representation of the whole image may not be corrected at these puny efforts. 
Secondly, exploring a fine-grid feature representation has been demonstrated as an efficient strategy to achieve an advanced ReID framework~\cite{sun2018beyond,wang2018learning,zheng2019pyramidal}. However, occluded person images usually lack several important parts due to the obstacles. Under this condition, those invalid noises will easily provide an ill similarity with similar obstacles and induce an error result. 
To this end, two typical frameworks have been proposed to tackle the above issues. One of the mainstream frameworks~\cite{wang2020high,hou2021feature} aims to aggregate the information from the whole image and handles the above issue by compensating the invisible body regions by its visible near-neighbor. With the assistance of well-trained human parsing or body key point estimation networks, these methods can easily conduct a topology graph based on the body key point. By passing the information from visible node to invisible node, the influence of occluded regions will largely be alleviated. 
Although retained the information from the whole image, not only the information from visible near-neighbor may not be so convincing enough, but also the inevitable domain gap between the assistant pre-trained model and the ReID datasets highly increases the difficulty to obtain an effective and efficient model. 
Alongside the above strategy, discovering and aligning the fine-grid visible body part in the spatial level of the occluded person image are a prevailing and straightforward strategy that has received much attention~\cite{qian2018pose,miao2019pose,gao2020pose}. By ignoring those invisible parts, these works achieve an significant improvement and visualization result. Nevertheless, to better distinguish the visible/invisible regions, most of the methods in this body of work also rely on extra pre-trained networks to provide body clues and suffer the same domain gap. To make matters worse, those error segmentation or key point results will make the valid nuances be abandoned easily.

Therefore, in this paper, we propose a Dynamic Prototype Mask (DPM) which not only escapes from the extra pre-trained networks but also simultaneously retains the information from the whole image and achieves alignment. The DPM is conducted under two self-evident prior knowledge: 1) During the training, the loss scale of high-quality images which suffer less occlusion will be lower than those of high-occluded images. Since the fully connected layer used for classification could be considered as the bank of prototype for each class, under this perspective, the lower loss scale can be seen as a high similarity between the high-quality sample and its corresponding prototype. In other words, those prototypes for each identity can be considered as a high-quality and complete feature representation that suffers little occlusion. 2) Each channel in the feature representation can be regarded as a response to a specific pattern. This phenomenon in CNN has already been explored by several previous works~\cite{chen2017sca,hu2018squeeze}. For the transformer, the multi-head self-attention directly aggregates the feature based on the similarity from patch to patch. These two prior knowledge indicate that the alignment and matching in the training period for occlusion person re-identification can largely be addressed by selecting a visible pattern subspace for both the input image and its holistic prototype. Motivated by this observation, we introduce the DPM. Different from the spatial attention strategy~\cite{chen2021occlude} which takes effect on the feature of the input image itself, the DPM aims to learn a dynamic mask to cut the holistic prototype and select the efficient subspace for matching. Meanwhile, this processing is totally spontaneous and does not rely on any extra network to provide body clues.

\begin{figure*}[t]
\centering
\includegraphics[width=2.1\columnwidth]{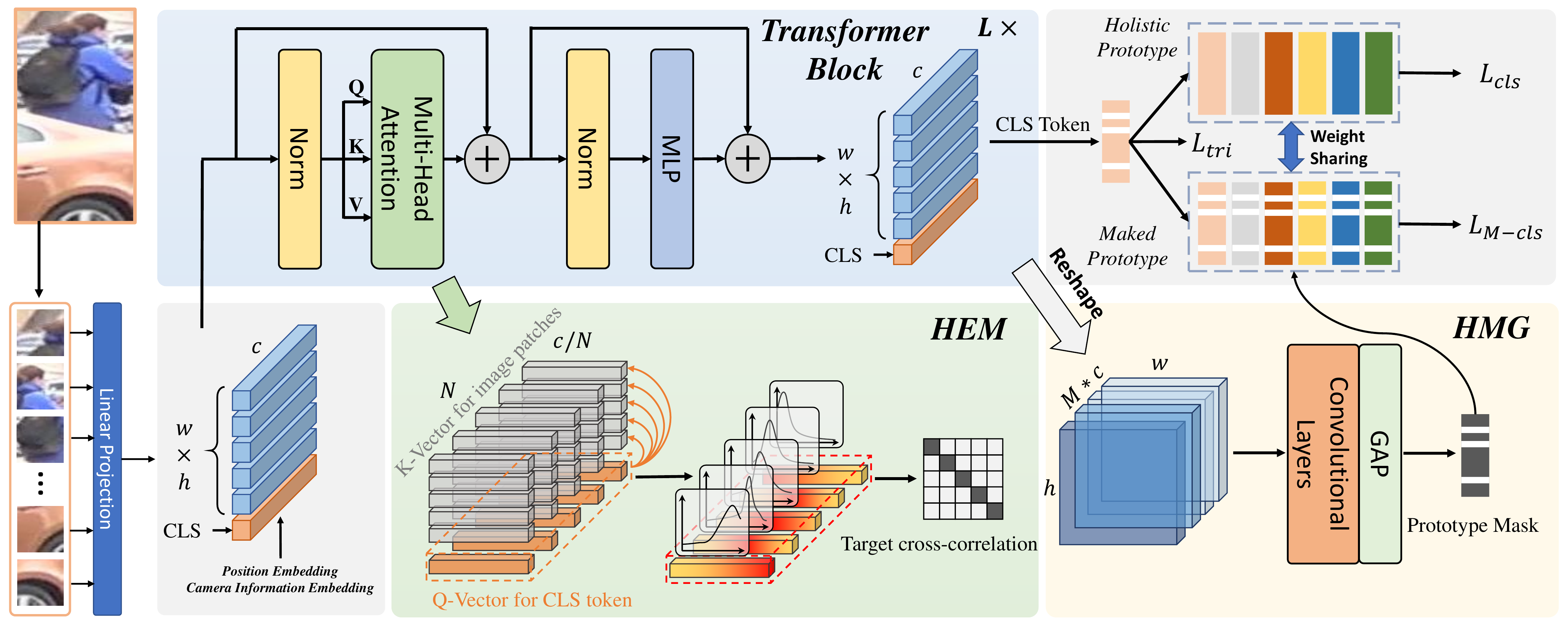}
\caption{The framework of proposed Dynamic Prototype Mask (DPM). Here, "HEM" denotes the Head Enrich Module, the "HMG" refers to the Hierarchical Mask Generator (HMG). $N$ is the number of the head in transformer blcok. $M$ is the number of feature maps that be concatenated by HMG. For the input occluded images, HEM encourage the multiple heads in the transformer block to aggregate pattern from different patches. Subsequently, a large amount of enriched feature representation of the dataset will be used to train a holistic prototype for each identity. For each single image with its occluded feature representation,  the HMG will provide a dynamic mask to select the appropriate subspace from the holistic prototype.}
\label{fig:overview}
\end{figure*}

To do this, as shown in Figure~~\ref{fig:overview}, the DPM starts with a standard ViT~\cite{dosovitskiy2020image} which has demonstrated its superior performance in the computer vision tasks before~\cite{chen2022lctr,luo2022towards,he2021transreid,li2021diverse}. Since the key idea of DPM is to generate the prototype mask to select the visible subspace for matching, we first introduce a hierarchical mask generator (HMG) to provide a reliable mask feature. The HMG takes the advantage of the convolutional neural network and aims to evaluate the weight for each channel by the correlation of local information. Meanwhile, we observe that with the network going deeper, the feature representation of each patch will be smoothed and become more similar to each other. Based on high-similar input, it is difficult for pure HMG to provide an efficient prototype mask. Therefore, we add a hierarchical structure to enhance the diversity of the input feature by shallow layers with high diversity. To fully explore the potential of DPM, an Head Enrich Module (HEM) is devised to enrich the feature representation. Specifically, each head in the final transformer block will be encouraged to aggregate different patterns in the whole image. Finally, to evaluate the effectiveness of the proposed DPM, we conduct a series of experiments on both occluded and holistic ReID benchmarks. 

The main contributions of the paper are summarized as follows:
\begin{itemize}
    \item A novel end-to-end trainable network DPM is proposed. DPM not only escapes from the extra pre-trained networks but also simultaneously retains the information from the whole image and achieves automatic alignment. 
    
    \item To fully explore the potential of DPM, a Hierarchical Mask Generator (HMG) together with a Head Enrich Module (HEM) is introduced. The HMG provides a high-quality sub-space mask via hierarchical semantic information, while the HEM enriches the holistic prototype via diverse heads.
    
    \item Extensive experiments on two publicly occluded datasets Occluded-Duke and Occluded-REID demonstrate the superiority of our DPM.
\end{itemize}

\section{Related Works}
\subsection{Holistic Person Re-Identification} 
Holistic person re-identification aims to address the problem of matching people over a distributed set of non-overlapping cameras. The prior works mainly focus on exploring the hand-craft descriptors~\cite{ma2014covariance,yang2014salient,liao2015person} with a well-designed metric learning strategy~\cite{zheng2012reidentification,koestinger2012large}. With the resurge of deep learning, deep feature representation learning has dominated the 
vision tasks \cite{he2017mask,mou2020plugnet,chen2021e2net,peng2021knowledge,zhang2021towards}. 
Luo \emph{et al}.~\cite{luo2019bag} introduce the BN-Neck structure in the CNN-based ReID framework. The research provides a strong baseline for the holistic ReID.
Chen \emph{et al}.~\cite{Chen_2019_Mixed} introduced a high-order attention mechanism to capture and use high-order attention distributions. 
Zheng \emph{et al}.~\cite{zheng2019joint} integrate discriminative and generative learning in a single unified network for person re-identification.
Besides using the global feature representation~\cite{luo2019bag,zheng2019joint,ye2021deep}, employing part-level features
for pedestrian image description to offer fine-grained information 
is also a mainstream strategy that has been verified as beneficial for person ReID. Methods like PCB~\cite{sun2018beyond}, MGN~\cite{wang2018learning}, and Pyramid~\cite{zheng2019pyramidal} horizontally divide the input images or feature maps into several parts to conduct a fine-grid representation. Most recently, we have witnessed the thriving of transformer structures from natural language processing to computer vision. TransReID~\cite{he2021transreid} firstly takes the advantage of ViT structure and applies it to the ReID task. Although those methods reach a satisfying performance in the holistic ReID benchmarks, the widely existing occlusion condition is largely ignored. Most of the methods suffer significant performance degradation when being applied to the real-world scenarios which contain the occluded cases.

\subsection{Occluded Person Re-Identification}  
Occluded person re-identification points out the weakness of holistic ReID methods in such occluded cases. The main challenge of occluded ReID lies in the incomplete body information which can not provide high-quality feature representation. To tackle this issue, early works attempt to remove the influence of obstacles in an end-to-end framework and generate the global feature representation from the visible part. Zhuo \emph{et al}.~\cite{zhuo2018occluded} introduce an extra occluded/non-occluded binary classification task to distinguish the occluded images from holistic ones. Chen \emph{et al}.~\cite{chen2021occlude} combine an occlusion augmentation scheme with an attention mechanism to precisely capture body parts regardless of the occlusion. Although this kind of work is easy to achieve and shows a good performance before, it always suffer the noise caused by the obstacles which limited the performance upper bound. Therefore, recent methods attempt to avoid such a condition with two typical strategies. The first one aims to aggregate the information from the whole image and handles the above issue by compensating for the invisible body regions by its visible near-neighbor. Wang \emph{et al}.~\cite{wang2020high} utilize the high-order relation and human-topology information that is based on keypoint estimation to learn well and robustly aligned features. Hou \emph{et al}.~\cite{hou2021feature} propose a region feature completion module to exploit the long-range spatial contexts from non-occluded regions to predict the features of occluded regions. Though aggregating the information from the visible neighbor node can alleviate the occluded condition, this process is still facing a great challenge when lacking efficient evidence in the neighbor nodes. Meanwhile, using the key point estimate network which is pre-trained on other datasets also faces a  challenge to provide reliable results when suffering a domain variation. Another strategy inherits the idea of fine-grid feature representation and aims to match the image between the visible parts. Miao \emph{et al}.~\cite{miao2019pose} introduce the  Pose-Guided Feature Alignment (PGFA), exploiting pose landmarks to disentangle visible part information from occlusion noise. Gao \emph{et al}.~\cite{gao2020pose} introduce the Visible Part Matching (PVPM) model to learn discriminative part features via a pose-guided attention map. Li \emph{et al}.~\cite{li2021diverse} employ the prototypes to disentangle the fine-grid body part without the help of an extra network in order to achieve satisfying performance. However, most of the fine-grid methods still rely on the extra network to provide body clues and suffer the same domain variation problem. Furthermore, since the fine-grid methods demand strict part prediction to send the feature to its corresponding branch, those incorrect results will  make those valuable nuances be ignored easily.

Differing from the above methods, the DPM not only escapes from the extra pre-trained networks but also simultaneously retains the global information from the whole image representation and achieves an automatic alignment.

\section{The Proposed Method}
\subsection{Overall Framework}
The overview of our proposed DPM framework is illustrated in Figure~~\ref{fig:overview}. The DPM adopts a pre-trained ViT~\cite{dosovitskiy2020image} to extract the original feature representation from the input images. Herein, we denote the input image as $I$ with the resolution as $H\times W$, We first split the image into $D (h \times w)$ patches with the size after flatten as $x_i \in 1\times c$. Specifically, it can be described as:
\begin{equation}
D = w \times h = \left \lfloor \frac{H-P+s_d}{s_{d}} + \frac{W-P+s_{d}}{s_{d}} \right \rfloor,
\end{equation}
where the $P$ and $s_{d}$ refers to the size of image patch and the step size of sliding window. After the linear projection $\mathcal{F}$, a learnable class token $x_{cls}$ is attached to aggregate the information from image patches. Before feeding into the transformer block, following the TransReID~\cite{he2021transreid}, a learnable position embedding $\mathcal{P}$ and camera embedding $\mathcal{C}$ is added to the patch embeddings to retain positional information and camera information respectively, which can be formulated as:
\begin{equation}
z_0 = [x^0_{cls};\mathcal{F} (x^0_1);\mathcal{F}(x^0_2);\cdots;\mathcal{F}(x^0_D)] + \mathcal{P} + \lambda \mathcal{C},
\end{equation}
where the $z_0$ is the input of the transformer blocks. The hyper-parameter $\lambda$ is used to balance the weight of camera embedding. The query and key vectors of the last transformer block are fed into the head enrich module (HEM), which takes the advantage of the multi-attention structure to explore a diverse feature representation for the different heads. Meanwhile, the representation of class-token will be utilized to train a holistic prototype for each class. To well tackle the occluded case, the representation for image patches of the 2$_{st}$, 4$_{th}$, 10$_{th}$, and 12$_{th}$ is concatenated and sent to the hierarchical mask generator (HMG) to provide the dynamic prototype mask for every single input image. Different from spatial attention methods~\cite{chen2021occlude} which take effect on the feature map itself, the prototype mask is used to cut the holistic prototype to select the subspace of high-discriminative visible patterns.

\begin{figure}[t]
    \centering
    \includegraphics[width=0.90\columnwidth]{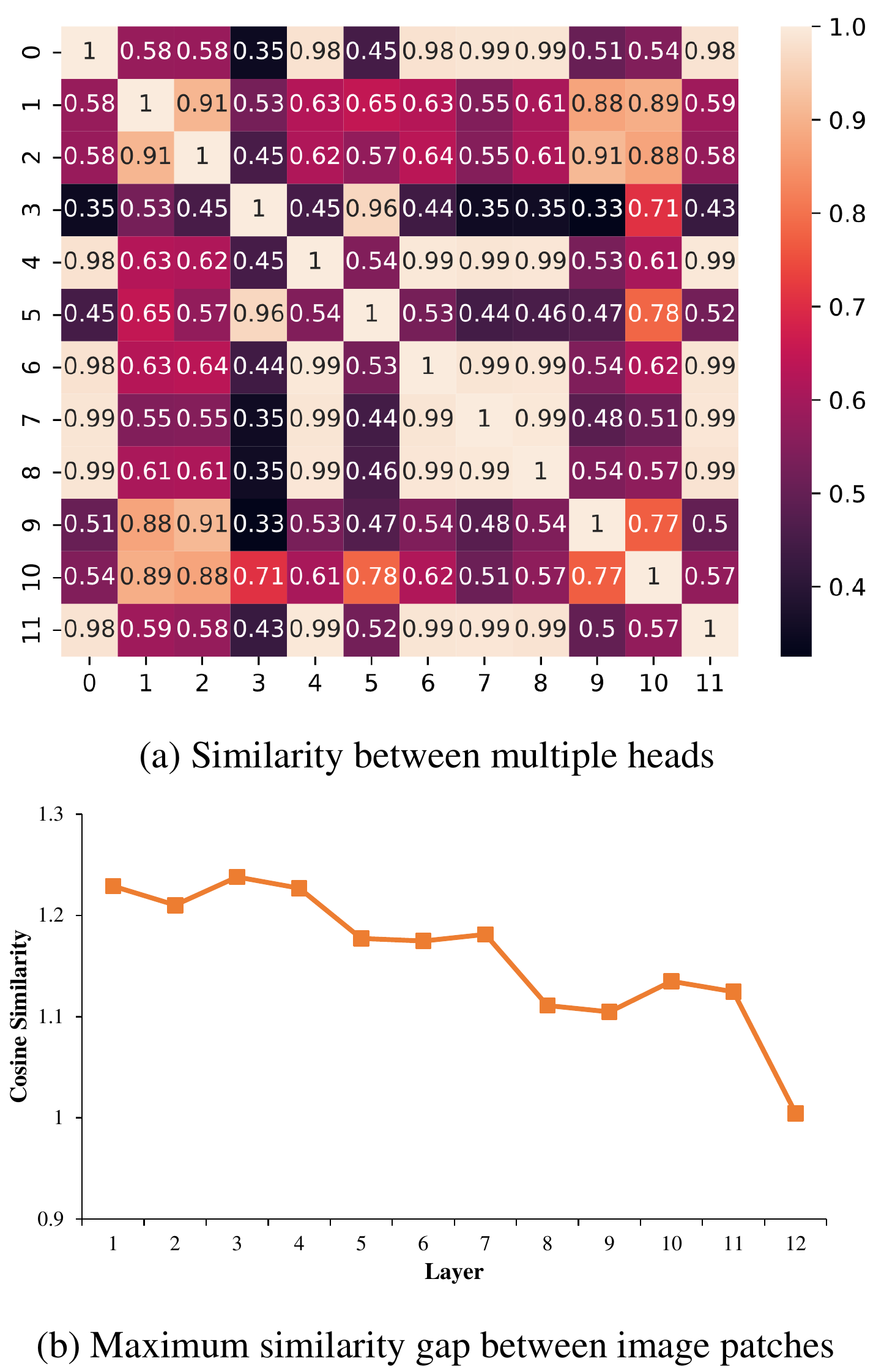}
    \caption{The motivation for Hierarchical Mask Generator (HMG) and Head Enrich Module (HEM). (a) The cross-correlation matrix between multiple heads' attention maps in the last transformer block. Several class tokens aggregate features from similar image patches which limit the diversity of representation. (b) The maximum similarity gap in each block. Herein, we have ignored the class token. Clearly, the feature of image patches has been smoothed and become more similar to each other with the network going deeper.}
    \label{fig:example}
\end{figure}

\subsection{Hierarchical Mask Generator}
Based on the two prior knowledge mentioned before, the main idea of DPM is to explore a spontaneous alignment and select a discriminative subspace for the holistic prototype to match every single image. Therefore, one of the most important parts for achieving the DPM is to generate an efficient prototype mask. In most cases, a pattern within an image is usually conducted by pixels that are spatially concentrated and form a connected component. Therefore, using a local region-based sliding window to weigh the importance of patterns is quite suitable in such an application. In order to provide a high-quality prototype mask, as shown in Figure~\ref{fig:overview}, we apply a convolutional-based mask generator that can take the neighbor nodes into consideration for each patch. 

In specific, after the $l_{th}$ block, HMG adopts the reshaped image representation $f_l \in \mathbb{R}^{h \times w \times c} $ by excluding the class-token in the feature representation $z_l$. Although directly using the image representation provided by the last block seems like the most intuitive strategy, as shown in Figure~\ref{fig:example} (b), after calculating the cosine similarity between the most dissimilar image patches in every block, we observe that by passing information through the similarity among image patches, the feature representations will be smoothed and become more similar. 
Those representations with few discriminative make it hard to provide an efficient prototype mask. To this end, a hierarchical structure is utilized to aggregate those image representations from the shallow layers. Inspired by the success of Swin-Transformer~\cite{liu2021swin} which merges the patches after the $2_{nd}$, $4_{th}$, $10_{th}$ transformer block, we also pick the image representations from these three blocks and combine them with the last block as the input for the HMG. The final prototype mask is generated as:
\begin{equation}
\begin{split}
& M_p = \sigma (Avgpool(\mathcal{G}(G * (f_{1}, f_{2},\dots, f_{L})))), \\
& with \qquad f_l = Reshape[x^l_1;x^l_2;\cdots;x^l_D].
\end{split}
\end{equation}
Herein, the $\sigma$ refers to the sigmoid function. The $\mathcal{G}$ denotes the convolutional layers of HMG and the $G \in \mathbb{R}^{1 \times L}$ is a binary gate to choose the input for HMG.

Finally, the masked prototype $W_{mp}$ for $i_{th}$ input image is generated by Hadamard product of the row-extended prototype mask $M^i_p$ and the prototype weight matrix $W_p$ as:
\begin{equation}
W^i_{mp} = W_p \odot M^i_p.
\end{equation}

\subsection{Head Enrich Module}
Multi-head self-attention which aims to extract difference and discriminative feature representation is considered as the key component to conducting the transformer block. This kind of aggregation of different patterns suits the DPM both in training a complete holistic prototype matrix and selecting an aligned and discriminative subspace. However, as shown in Figure~\ref{fig:example} (a), in the original transformer block, there is no explicit optimization to encourage multiple heads to aggregate more nuances in the whole image, which makes it possible for the different heads to have similar feature embedding. The similar head representations will in turn limit the DPM to select a well-aligned subspace.

On account of this condition, we introduce a head enrich module (HEM) to push multiple heads in the class token to obtain diverse patterns in the last transformer block. Generally, the multi-head attention adopts the query matrix Q, key matrix K, and value matrix V to pass the information from different patches. In the training phase, we only employ the class token as the global representation. Therefore, when ignoring the key vector of the class token itself, we can obtain the attention map of $A \in \mathbb{R}^{N \times D}$ between the class token and the image patches as:
\begin{equation}
A(q^{cls}_{L},K^{img}_{L})=softmax(\frac{q^{cls}_{L} (K^{img}_{L})^T}{\sqrt{c / N}}),
\end{equation}
where $N$ is the number of heads, $q^{cls}_{L}$ is the query vector of class token in the $L_{th}$ transformer block, $K^{img}_{L}$ is the query vector of image patches in the $L_{th}$ transformer block, and $N$ refers to the number of heads. To push the attention map of each head apart, an orthonormal constraint is impose as:

\begin{equation}
\mathcal{L}_{hem} = \left \| \dot A \dot A^T - \mathbb{I}_N  \right \| ^2_F,
\end{equation}
where the $\left \| \cdot   \right \| ^2_F$ is Frobenius norm, the $\mathbb{I}_N \in N \times N$ is the identity matrix, $\dot A$ is a normalized $A$ matrix with each row being L2 normalized. With $\mathcal{L}_{hem}$, the class token could provide richer representation, which not only benefit the learning for holistic prototype and the masked generator.

\subsection{Loss Function and Optimization}
\label{sec:optimization}
The DPM contains a two-branch learning framework as the original classification loss and the masked classification loss. Intuitionally, we should employ the softmax loss to optimize both two branches. However, this strategy will cause a condition that the mask can not be well learned. After adding the mask, the scale of the loss generated by the masked branch is much smaller than the softmax loss, which can not provide enough power to learn a high-quality prototype mask. Although increasing the weight of the masked branch may make sense, the softmax loss is not a strong constraint to clustering the samples. Limited constrains also limits the performance of DPM. Inspired by the progress of metric learning~\cite{deng2019arcface}, we employ an extra angular margin in the original softmax loss to optimize the masked branch. This strategy not only balances the scale of the original branch and masked branch but also highlights the importance of learning a high-quality mask during the training phase. With the extra margin, for input $x_cls^{L}$ with label $y^i$, the $L_{M-cls}$ can be given as:
\begin{equation}
\begin{split}
& \mathcal{L}_{M-cls} = -\frac{1}{B} \sum_{i=1}^{B} log\frac{e^{s(cos(<x_{cls}^{L}, W^{y^i}_{mp}> + m))}}{e^{s(cos(<x_{cls}^{L}, W^{y^i}_{mp}> + m))}+ D_{inter}}  , \\
& with \qquad D_{inter} = {\textstyle \sum_{j=1, j \ne y^i}^{C}e^{s(cos(<x_{cls}^{L}, W^{y^i}_{mp}>))}},
\end{split}
\end{equation}
where the $B$ and $C$ refers to the batch size and the number of class. The $m$ denotes the angular margin and $s$ is the hyper-parameter to adjust the scale. 

To increase the intra-class similarity and decrease the inter-class similarity, triplet loss $\mathcal{L}_{tri}$ with online hard-mining~\cite{schroff2015facenet} are combined during the supervise training. Therefore, the overall loss function can be formulated as:
\begin{equation}
\mathcal{L} = \alpha \mathcal{L}_{cls} + (1 - \alpha) \mathcal{L}_{M-cls} + \beta \mathcal{L}_{hem} + \mathcal{L}_{tri}
\label{eq:overall}
\end{equation}
where the $\alpha$ and $\beta$ is the hyper-parameters to adjust the weight of $\mathcal{L}_{M-cls}$ and $\mathcal{L}_{hem}$ respectively.

\section{Experiment}
\subsection{Datasets and Experimental Setting}
\textbf{Datasets.} To evaluate the effectiveness of the proposed DPM, we conduct extensive experiments on four publicly available ReID benchmarks which include both occluded and holistic person re-identification datasets. The details are as follows.

\textbf{Occluded-Duke}~\cite{miao2019pose} is a large-scale dataset collected from the DukeMTMC for occluded person re-identification. The training set consists of 15,618 images of 702 persons. The testing set contains 2,210 images of 519 persons as the query and 17,661 images of 1,110 persons as the gallery. Until now, Occluded-Duke is still the most challenging dataset for occluded ReID due to its scale.

\textbf{Occluded-REID}~\cite{zhuo2018occluded} is an occluded person dataset captured by mobile cameras. It consists of 2000 images from 200 persons, where each person has 5 whole-body images and 5 occluded person images. Following the evaluation protocol of previous works~\cite{gao2020pose,wang2020high,chen2021occlude}, the Occluded-REID is only used as a testing set. The model used for experiments in this dataset is trained under the training set of Marker-1501~\cite{zheng2015scalable}.

\textbf{Market-1501}~\cite{zheng2015scalable} is a widely-used holistic ReID dataset captured from 6 cameras. It includes 12,936 training images of 751 persons as the training set, 3,368 images of 750 persons as the query, and 19,732 images of 750 persons as the gallery. 

\textbf{DukeMTMC-reID}~\cite{zheng2017unlabeled} contains 36,441 images of 1,812 persons captured by eight cameras, in which 16,522 images of 702 identities are used as the training set, 2,228 and 16,522 images of 702 persons that do not appear in the training set are used as the query and gallery, respectively.

\textbf{Evaluation Protocol.} To verify fair comparison with other methods, we adopt the widely used Cumulative Matching Characteristic (CMC) and mean Average Precision (\emph{m}AP) as evaluation metrics and follow the evaluation settings provided by existing occluded methods~\cite{wang2020high, gao2020pose}.

\textbf{Implementation details.}
We employ the ViT~\cite{dosovitskiy2020image} pre-trained on ImageNet~\cite{deng2009imagenet} as the backbone network. Particularly, we resize all the input images to $256 \times 128$ and adopt commonly used horizontal flipping, padding, random cropping, and random erasing~\cite{zhong2020random} as data augmentation. Following~\cite{wang2020high,yang2021learning}, we use extra color jitter augmentation to avoid domain variance when conduct testing in the Occluded-REID. Following the success of TransReID~\cite{he2021transreid}, we adopt a lower stride and set $\lambda$ to $3.0$.
During the training stage, each mini-batch is conducted by 64 images from 4 identities. In order to strengthen the power of HMG, the training phase is divided into two-step in every iteration. In the first step, we froze the parameter of HMG to train the holistic prototype. In the second step, we froze the parameter except the HMG to train a high-quality prototype mask. During the testing phase, we apply the mask generated by the query image to the gallery images. Then the retrieval stage can still be computed in parallel for each query image.
The SGD is utilized as the optimizer, in which the learning rate is initiated as $0.008$ with cosine learning rate decay. The hyper-parameters for $s$ and $m$ in Arcface loss are set to $30$ and $0.5$ respectively in training the Occluded-Duke. In training Occluded-REID, since the strong constraints will induce the overfitting easily when considering the domain variance of Market-1501 and Occluded-REID, we decrease the $m$ to $0.3$ and $\beta$ to $0.01$ in training. We implement our DPM with PyTorch and conduct all experiments on a single Nvidia Tesla A100.

\begin{table}[t]
  \centering
  \renewcommand{\arraystretch}{1.1}
  \resizebox{85mm}{!}{
  \begin{tabular}{l|cc|cc}
  \toprule[1pt]
   \multirow{2}{*}{\textbf{Method}} &\multicolumn{2}{c|}{\textbf{Occluded-Duke}} & \multicolumn{2}{c}{\textbf{Occluded-REID}}\\
   \cline{2-5}
   & R-1 & \emph{m}AP & R-1 & \emph{m}AP\\
  \hline
    PCB \cite{sun2018beyond} & 42.6 & 33.7 & 41.3 & 38.9 \\
    Part Bilinear \cite{suh2018part}           & 36.9 & - & - & - \\
    FD-GAN \cite{ge2018fd}                 & 40.8 & - & - & - \\
    ISP \cite{zhu2020identity}                    & 62.8 & 52.3 & - & -\\
    TransReID*  \cite{he2021transreid}             & 66.4 & 59.2 & - & -\\
    \hline    
    DSR \cite{he2018deep}                    & 40.8 & 30.4 & 72.8 & 62.8\\
    Ad-Occluded \cite{huang2018adversarially}             & 44.5 & 32.2 & - & -\\
    FPR \cite{he2019foreground}                     & - & - & 78.3 & 68.0\\
    PGFA \cite{miao2019pose}                   & 51.4 & 37.3 & - & -\\
    PVPM+Aug \cite{gao2020pose}                & - & - & 70.4 & 61.2\\
    HOReID \cite{wang2020high}                 & 55.1 & 43.8 & 80.3 & 70.2\\
    OAMN \cite{chen2021occlude}                   & 62.6 & 46.1 & - & -\\
    Part-Label \cite{yang2021learning}              & 62.2 & 46.3 & 81.0 & 71.0\\
    PAT* \cite{li2021diverse}                     & 64.5 & 53.6 & 81.6 & 72.1\\
  \hline
  \textbf{DPM}              & \textbf{71.4} & \textbf{61.8} & \textbf{85.5} & \textbf{79.7}\\
  \bottomrule[1pt]
    \end{tabular}}
    \vspace{0em}\caption{\textbf{Comparision with previous state-of-the-art methods in terms of CMC (\%) and \emph{m}AP (\%) on Occluded-Duke and Occluded-REID.} 
    The symbol $*$ represents methods that employ the transformer structure.}
    \label{Tbl:osota}
    \vspace{-2em}
\end{table} 

\begin{table}[t]
  \centering
  \renewcommand{\arraystretch}{1.1}
  \resizebox{85mm}{!}{
  \begin{tabular}{l|cc|cc}
  \toprule[1pt]
   \multirow{2}{*}{\textbf{Method}} &\multicolumn{2}{c|}{\textbf{Market-1501}} & \multicolumn{2}{c}{\textbf{DukeMTMC}}\\
   \cline{2-5}
   & R-1 & \emph{m}A & R-1 & \emph{m}AP\\
   \hline
    PCB \cite{sun2018beyond}         & 92.3 & 71.4 & 81.8 & 66.1 \\
    MGN \cite{wang2018learning}                      & \textbf{95.7} & 86.9 & 88.7 & 78.4 \\
    ISP \cite{zhu2020identity}       & 95.3 & 88.6 & 89.6 & 80.0\\
    CDNet \cite{li2021combined}                    & 95.1 & 86.0 & 88.6 & 76.8\\
    TransReID* \cite{he2021transreid} & 95.2 & 88.9 & 90.7 & 82.0\\
   \hline
    FPR \cite{he2019foreground}      & 95.4 & 86.6 & 88.6 & 78.4\\
    PGFA \cite{miao2019pose}         & 91.2 & 76.8 & 82.6 & 65.5\\
    HOReID \cite{wang2020high}       & 94.2 & 84.9 & 86.9 & 75.6\\
    OAMN \cite{chen2021occlude}      & 93.2 & 79.8 & 86.3 & 72.6\\
    PAT* \cite{li2021diverse}         & 95.4 & 88.0 & 88.8 & 78.2\\
   \hline
    \textbf{DPM}              & 95.5 & \textbf{89.7} & \textbf{91.0} & \textbf{82.6}\\
   \bottomrule[1pt]
  \end{tabular}}
  \vspace{0em}\caption{\textbf{Comparision with state-of-the-art methods in terms of CMC (\%) and \emph{m}AP (\%) on Market-1501 and DukeMTMC-reID.} 
  The symbol $*$ represents methods that employ the transformer structure.}
  \label{Tbl:hsota}
  \vspace{-2em}
\end{table} 

\subsection{Comparison with State-of-the-art Methods}
\textbf{Results on Occluded Datasets.}
To comprehensively demonstrate the performance of DPM, we evaluate DPM against the previously reported state-of-the-art methods on the Occluded-Duke and Occluded-REID in Table~\ref{Tbl:osota}, The compared methods include holistic ReID methods~\cite{sun2018beyond,suh2018part,ge2018fd,zhu2020identity,he2021transreid} and occluded ReID methods~\cite{he2018deep,huang2018adversarially,he2019foreground,miao2019pose,gao2020pose,wang2020high,li2021diverse,chen2021occlude,yang2021learning}. Obviously, the transformer-based structure (PAT, TransReID) has the advantage in solving occluded cases when compared to the convolutional neural network. The DPM also inherits this advantage and further outperforms other transformer-based methods. In the most challenging occluded ReID dataset Occluded-Duke, the DPM reaches an impressive performance, with $71.4\%$ in rank-1 and $61.8\%$ in \emph{m}AP, which at least outperforms other occluded ReID methods with $6.9\%$ and $8.2\%$ in rank-1 and \emph{m}AP respectively. Meanwhile, in the Occluded-REID, the proposed DPM consistently surpasses current state-of-the-art methods. Specifically, the DPM achieves $85.5\%$ in rank-1 accuracy  \and $79.7\%$ in \emph{m}AP, which improves the Rank-1 accuracy by $3.9\%$ and \emph{m}AP by $7.6\%$ over the PAT.

Although not relying on the extra network to provide body clues, the DPM still achieves superior performance in the occluded ReID benchmarks. 

\textbf{Results on Holistic Datasets.}
Although occluded ReID methods mainly focused on solving the occluded ReID issue, they may suffer a performance decrease in the original holistic ReID task due to incorrect alignment or ignoring of valuable regions. Therefore, in this section, we also evaluate the proposed DPM on the holistic ReID dataset Market-1501 and DukeMTMC-ReID. For better comparison, we select five holistic ReID method~\cite{sun2018beyond,wang2018learning,zhu2020identity,li2021combined,he2021transreid} and five occluded ReID methods~\cite{he2019foreground,miao2019pose,wang2020high,chen2021occlude,li2021diverse}. 

The results are shown in Figure~\ref{Tbl:hsota}. In the Market-1501 dataset, the DPM gets $95.5\%$ in rank-1 accuracy and $89.7\%$ in \emph{m}AP. In the DukeMTMC-reID, the  the DPM gets $91.0\%$ in rank-1 accuracy and $82.6\%$ in \emph{m}AP.
It is clear that DPM shows competitive results in both two holistic ReID datasets when compared to the state-of-the-art holistic ReID methods. When compared to the occluded ReID methods, the DPM outperforms the previous state-of-the-art methods in these two datasets. Overall, the above results show that DPM is a universal framework, which mainly aims to tackle occluded cases. But it will not destroy the performance on the general holistic ReID task.  

\begin{table}[t]
  \centering
  \renewcommand{\arraystretch}{1.1}
  \resizebox{80mm}{!}{
  \begin{tabular}{l|cccc}
  \toprule[1pt]
   \multirow{2}{*}{\textbf{Method}} & \multicolumn{4}{c}{\textbf{Occluded-Duke}} \\
   \cline{2-5}
   & R-1 & R-5 & R-10 & \emph{m}AP\\
  \hline
    baseline            & 64.8 & 80.8 & 85.8 & 57.8 \\
    +DPM                & 70.1 & 82.8 & 86.9 & 59.9 \\
    +DPM+HR             & 71.0 & 82.9 & 87.2 & 61.0 \\
    +DPM+HR+HEM         & \textbf{71.4} & \textbf{83.7} & \textbf{87.4} & \textbf{61.8}\\
  \bottomrule[1pt]
    \end{tabular}}
    \vspace{0em}\caption{\textbf{Ablation study of each components in DPM on Occluded-Duke.} Herein, the HR refers to the hierarchical structure in mask generator, the HEM referes to the head enrich module. }
    \label{Tbl:ablation}
    \vspace{-1em}
\end{table} 

\begin{table}[t]
  \centering
  \renewcommand{\arraystretch}{1.1}
  \resizebox{85mm}{!}{
  \begin{tabular}{l|cc|cccc}
  \toprule[1pt]
   \multirow{2}{*}{\textbf{Method}} & \multicolumn{2}{c|}{\textbf{Setting}} & \multicolumn{4}{c}{\textbf{Occluded-Duke}} \\
   \cline{2-7}
   & Cls & Mask-Cls & R-1 & R-5 & R-10 & \emph{m}AP\\
  \hline
    baseline            &$\mathcal{S}$ &              & 64.8 & 80.8 & 85.8 & 57.8 \\
    baseline            &$\mathcal{A}$ &              & 68.6 & 81.2 & 85.9 & 58.5 \\
    DPM                 &$\mathcal{S}$ &$\mathcal{S}$ & 64.6 & 80.6 & 85.5 & 57.3 \\
    DPM                 &$\mathcal{A}$ &$\mathcal{A}$ & 68.3 & 80.0 & 84.2 & 57.7\\
    DPM                 &$\mathcal{S}$ &$\mathcal{A}$ & \textbf{70.1} & \textbf{82.8} & \textbf{86.9} & \textbf{59.9}\\
  \bottomrule[1pt]
    \end{tabular}}
    \vspace{0em}\caption{Performance comparison with different combinations of loss function for classifier and mask-classifier on Occluded-Duke. Here, the $\mathcal{S}$ denotes the branch which is training with the original softmax loss, and the $\mathcal{A}$ denotes the branch which is training with an extra angular margin.}
    \label{Tbl:DPM}
    \vspace{-1em}
\end{table}

\subsection{Ablation Study}
To evaluate the influence of the proposed architectural components. We conduct a series experiments over the occluded-Duke with different settings and show the quantitative results in Table~\ref{Tbl:ablation}. The baseline uses the ViT as backbone and training with the original softmax loss $L_{cls}$ and triplet loss $L_{tri}$.

Compared to the baseline method, adding the DPM strategy highly improved the performance in both rank-1 accuracy and \emph{m}AP as $1$ and $1$ respectively. After adding the hierarchical structure in the mask generator, the performance further increases from $1$ and $1$ to $1$ and $1$ in rank-1 and \emph{m}AP. On the other hand, it also demonstrates that the image representation provided by the last transformer block which lacks sufficient diversity information will limit the performance of mask generator. Meanwhile, HEM also provides significant performance gains as $1$ and $1$ in rank-1 and \emph{m}AP based on the above results. The experiments results indicate that all these components have made sense and satisfied their motivation in the DPM. All the components contribute to an effective framework consistently and finally result in an impressive performance.

\begin{figure}[t]
    \centering
    \includegraphics[width=0.90\columnwidth]{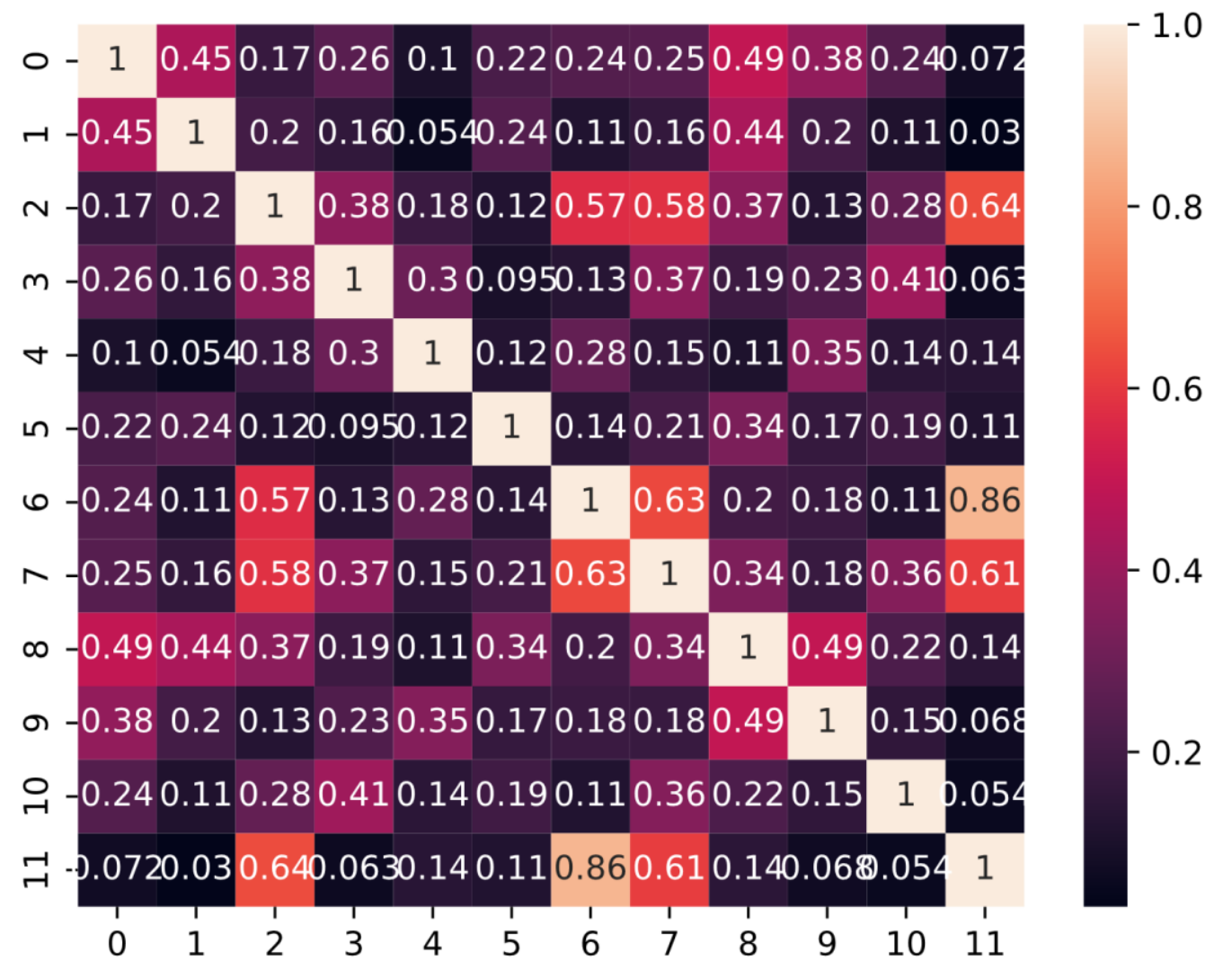}
    \caption{The cross-correlation matrix between multiple heads' attention maps in the class token of the last transformer block after adding the HEM. For better comparison, we visualize the same input image as the Figure~\ref{fig:example} (a).}
    \label{fig:hem}
\end{figure}

\begin{table}[t]
  \centering
  \renewcommand{\arraystretch}{1.1}
  \resizebox{80mm}{!}{
  \begin{tabular}{cccc|cccc}
  \toprule[1pt]
   \multicolumn{4}{c|}{\textbf{Setting}} & \multicolumn{4}{c}{\textbf{Occluded-Duke}} \\
   \cline{1-8}
   $P_{n}$ & $F_{n}$ & $P$ & $F$ & R-1 & R-5 & R-10 & \emph{m}AP\\
  \hline
    \checkmark &            &  &  & \textbf{71.4} & 83.7 & \textbf{87.4} & \textbf{61.8} \\
    \checkmark & \checkmark &  &  & 69.5 & 82.3 & 86.8 & 58.5 \\
      &  & \checkmark &            & 70.2 & \textbf{84.0} & \textbf{87.4} & 60.8 \\
     &  & \checkmark & \checkmark & 69.7 & 81.6 & 85.1 & 58.3\\
  \bottomrule[1pt]
    \end{tabular}}
    \vspace{0em}\caption{Performance comparison with different types of DPM in terms of CMC (\%) and \emph{m}AP (\%) on Occluded-Duke. Here, the $P$ and $F$ denotes that the mask is taking effect upon the prototype matrix and the feature representation respectively. The $P_{n}$ and $F_{n}$ denote that the mask is taking effect after the L2 normalization.}
    \label{Tbl:PF}
    \vspace{-2em}
\end{table}

\subsection{Discussions}
\textbf{The classification loss function for DPM.} As the most important module, how to train a efficient mask generator is one of the main challenges to achieve the the DPM. In Section~\ref{sec:optimization}, we have mentioned that we utilize an extra angular margin to train the masked branch to avoid inefficient optimization for the masked branch. Therefore, in this part, we conduct a comparison to show the performance of different loss function combinations when training the class branch and masked branch. The results are shown in Table~\ref{Tbl:DPM}. Here, we use the $\mathcal{S}$ to denote the branch which is training with the original softmax loss, and the $\mathcal{A}$ to denote the branch which is training with an extra angular margin. The two baselines are training without the masked branch.

From Table~\ref{Tbl:DPM}, we can observe that adding an extra angular margin could improve the rank-1 accuracy since it can enforce the network to pay more attention to those outlier samples. However, we can also observe that this strategy can not bring such a significant increment in the \emph{m}AP, which denotes that the whole distribution may not be ameliorated. 
Meanwhile, as we have mentioned before, adding an extra mask to select the subspace will further decrease the scale of the loss. It indicates that the mask generator just needs to output the same score for all the channels, the scale loss will still be satisfying. Therefore, as shown in Table~\ref{Tbl:DPM}, when using the same loss function in two branches, the optimization of the mask generator will be limited and the whole network will degrade to optimize the classification branch. After adding an extra angular margin in the masked branch and keeping the original classification branch, the mask generator has been encouraged to adopt a more radical optimization thus alleviating the above dilemma. As shown in the last line in Table~\ref{Tbl:DPM}, this kind of strategy obtain a significant improvement in both rank-1 accuracy and \emph{m}AP accuracy.

\begin{figure}[t]
    \centering
    \includegraphics[width=0.9\columnwidth]{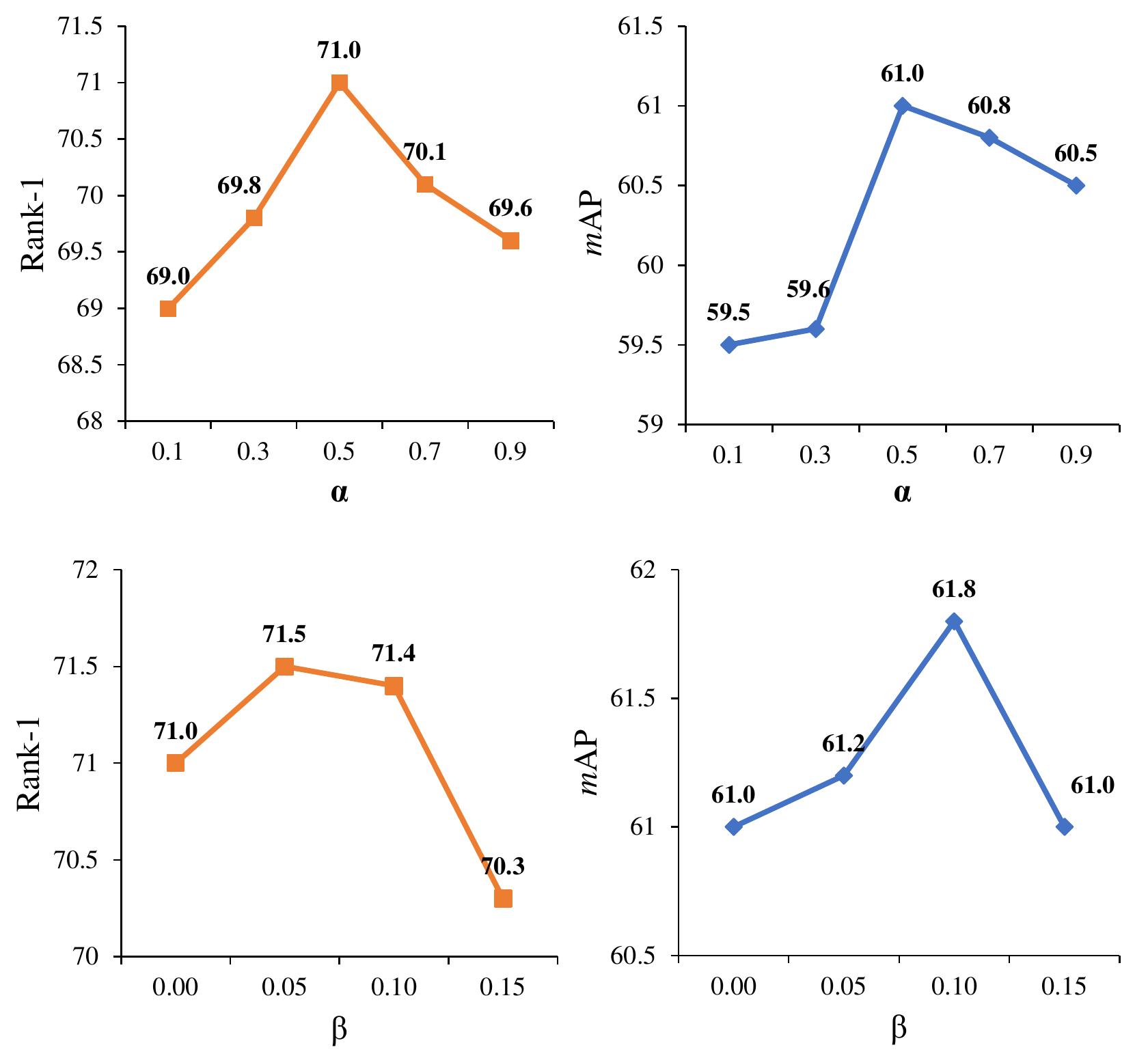}
    \vspace{-1em}
    \caption{\textbf{Impact of the hyper-parameters $\alpha$ and $\beta$ in terms of CMC (\%) and \emph{m}AP (\%) on Occluded-Duke.}  The performance reaches the peak when we set the $\alpha$ to $0.5$ and $\beta$ to $0.1$.}
    \label{fig:hyper}
\end{figure}

\textbf{Effect of head enrich module.} 
The HEM aims to enrich the feature representation of multiple heads in the class token to overcome the tendency that more different head has similar representations. The HEM can help not only the training for a holistic prototype with more nuances but also the training for mask generator to select the potential subspace. In Figure~\ref{fig:example} (a), we have visualized the cross-correlation matrix between multiple heads' attention map to show the limitation of the original transformer block. Therefore, in this part, we visualized the cross-correlation matrix with the same input once again in Figure~\ref{fig:hem} after adding the HEM. 

By adding the HEM that can explicitly encourage each head to have a different attention map, we could observe that the cross-correlation matrix between multiple heads' attention maps has significantly decreased as shown in Figure~\ref{fig:hem}. It indicates that the class token has received more diverse pattern information from the different image patches. Overall, the visualization well demonstrates the effectiveness of the HEM in such an application. 

\textbf{Impact of the hyper-parameters $\alpha$ and $\beta$.}
As indicated by the loss function in Eq. \ref{eq:overall}, we set two hyper-parameters $\alpha$ and $\beta$ to balance the weight of different components in the overall loss functions. Specifically, the $\alpha$ controls the trade-off between the holistic prototype matrix and the prototype mask, while the $\beta$ controls the correlation between different heads.
Hence, in this part, we conduct empirical experiments to measure the performance of the model under different hyper-parameters settings. When discussing the $\alpha$, we select the DPM with HMG as the baseline to conduct the experiments. As we have mentioned before the performance is sensitive to the $\alpha$, a small $\alpha$ will limit the ability to learn a holistic prototype matrix while a large $\alpha$ can not provide enough power to learn a high-quality prototype mask. As shown in the Figure~\ref{fig:hyper}, we observe that the performance of rank-1 accuracy and \emph{m}AP increase linearly when the $\alpha$ is less than $0.5$.  The performance reaches the peak when the $\alpha$ is set to $0.5$ with $71.0\%$ and $61.0\%$ in rank-1 accuracy and \emph{m}AP. After that, a larger $\alpha$ will decrease the performance.

Based on the model, we further discuss the influence of $\beta$. As shown in Figure~\ref{fig:hyper}, after adding the HEM into training, the rank-1 accuracy and \emph{m}AP also get improved when $\beta$ is less than $0.15$. The best performance in rank-1 reaches when the $\beta$ is set to $0.05$ and the best \emph{m}AP reaches when the $\beta$ is set to $0.10$. Considering the overall performance, we select the $\beta$ as $0.10$ in the further experiments.

\textbf{Different types of DPM.}
In DPM, we apply the mask to the holistic prototype matrix, while attention-based strategies explore the spatial attention mask which takes effect on the input image itself to alleviate the noise caused by obstacles. Therefore, in this part, we also make discuss whether the mask should also be applied to the feature representation. Meanwhile, in this part, we also evaluate the influence of L2 normalization in the DPM. Herein, we select the complete DPM as the baseline in the comparison and give an empirical analysis on the Occluded-Duke. The experimental results are shown in Table \ref{Tbl:PF}.

Benefits from the great ability of transformer structure that provides an effective representation for the visible part, in both settings, applying the mask upon the feature representation can not provide an extra performance gain. On another side, under both settings, applying the mask on the prototype after the L2 normalization works better than applying the mask before the L2 normalization. Thus, we select the only prototype mask which is applied before the L2 normalization as the final model.

\section{Conclusion}
In this paper, we address the occluded person re-identification with a novel dynamic prototype mask (DPM). The DPM takes the advantage of prototype classification and transfers the alignment in occluded retrieval to the subspace selection task. This strategy not only gets rid of the extra pre-trained networks to provide body clues but also simultaneously retains the information from the global wise and achieves an automatic alignment. Meanwhile, based on the observation in the original DPM framework, we further explore a Hierarchical Mask Generator (HMG) together with a Head Enrich Module (HEM) to fully exploit the potential of DPM. Finally, extensive experiments on occluded and holistic datasets demonstrate the superior performance of DPM.

\section*{Acknowledgments}
This work was supported by the National Science Fund for Distinguished Young Scholars (No.62025603), the National Natural Science Foundation of China (No.U1705262, No.62176222, No.62176223, No.62176226, No.62072386, No.62072387, No.62072389, No.62002305, No.61772443, No.61802324 and No.61702136), Guangdong Basic and Applied Basic Research Foundation (No.2019B1515120049), the Natural Science Foundation of Fujian Province of China (No.2021J01002), and the Fundamental Research Funds for the Central Universities (No.20720200077, No.20720200090 and No.20720200091).

\bibliographystyle{ACM-Reference-Format}
\bibliography{sample-base}

\end{document}